\documentclass[a4paper,conference]{IEEEtran}

\usepackage{graphicx}
\usepackage{amsmath}
\usepackage{amssymb}
\usepackage{multirow}
\usepackage{paralist}
\usepackage{booktabs}
\usepackage{cellspace}
\usepackage{subfig}
\usepackage{wrapfig}

\def\marrow{{\marginpar[\hfill$\longrightarrow$]{$\longleftarrow$}}}

\def\kobus #1 {{\textcolor{red}{ \sc \newline\newline Kobus says: }{\marrow\sf #1 \newline\newline}}}

\usepackage[pagebackref=true,breaklinks=true,colorlinks,bookmarks=false]{hyperref}
\usepackage[numbers, sort&compress]{natbib}

\usepackage{cleveref}  
\crefformat{table}{Table~#2#1#3}
\crefformat{section}{Section~#2#1#3}
\crefformat{subsection}{subsection~#2#1#3}
\crefformat{subsubsection}{Subsubsection~#2#1#3}
\crefformat{figure}{Fig.~#2#1#3}
\crefformat{subfigure}{Fig.~#2#1#3}
\crefformat{equation}{Eq.~(#2#1#3)}
\crefformat{algorithm}{Algorithm~#2#1#3}
\crefformat{appendix}{Appendix #2#1#3}
\title{Attention as Activation}

\begin{document}

\author{Yimian Dai$^{1}$ \quad\quad Stefan Oehmcke$^2$ \quad\quad Fabian Gieseke$^{2,3}$\quad\quad Yiquan Wu$^{1}$ \quad\quad Kobus Barnard$^{4}$\\ [0.1in]
\normalsize
$^1$College of Electronic and Information Engineering, Nanjing University of Aeronautics and Astronautics, Nanjing, China \\
Email: \{yimian.dai, nuaaimage\}@gmail.com\\
$^2$Department of Computer Science, University of Copenhagen, Copenhagen, Denmark\\
Email: \{stefan.oehmcke, fabian.gieseke\}@di.ku.dk\\
$^3$Department of Information Systems, University of M\"unster, M\"unster, Germany\\
Email: fabian.gieseke@uni-muenster.de\\
$^4$Department of Computer Science, University of Arizona, Tucson, AZ, USA\\
Email: kobus@cs.arizona.edu\\
}

\maketitle

\begin{abstract}
Activation functions and attention mechanisms are typically treated as having different purposes and have evolved differently.
However, both concepts can be formulated as a non-linear gating function. 
Inspired by their similarity, we propose a novel type of activation units called attentional activation~(ATAC) units as a unification of activation functions and attention mechanisms.
In particular, we propose a local channel attention module for the simultaneous non-linear activation and element-wise feature refinement, which locally aggregates point-wise cross-channel feature contexts.
By replacing the well-known rectified linear units by 
such ATAC units in convolutional networks, we can construct fully attentional networks that perform significantly better with a modest number of additional parameters.
We conducted detailed ablation studies on the ATAC units using several host networks with varying network depths to empirically verify the effectiveness and efficiency of the units.
Furthermore, we compared the performance of the ATAC units against existing activation functions as well as other attention mechanisms on the CIFAR-10, CIFAR-100, and ImageNet datasets.
Our experimental results show that networks constructed with the proposed ATAC units generally yield performance gains over their competitors given a comparable number of parameters.

\end{abstract}


\IEEEpeerreviewmaketitle


\section{Introduction}\label{sec:intro}

Key technological advances in deep learning include the recent development of advanced \emph{attention mechanisms}~\cite{CVPR18SENet} and \emph{activation functions}~\cite{ICML10ReLU}.
While both attention mechanisms and activation functions depict non-linear operations, they are generally treated as two different concepts and have been improved and extended into different directions over recent years.
The development of novel attention mechanisms has led to more sophisticated and computationally heavy network structures, which aim to capture long-range spatial interactions or global contexts~\cite{NIPS18GENet,ECCV18CBAM,ICCV19AugmentedConv}. 
In contrast, activation functions, no matter if designed by hand~\cite{NIPS17SELU} or via automatic search techniques~\cite{ICLRW18Swish}, have remained scalar and straightforward. 

Recently, a few learning-based approaches~\cite{ICCV15PReLU,CVPR18xUnit} have been proposed that augment the \emph{rectified linear unit}~(ReLU) with learnable parameters. 
However, such learning-based activation functions still suffer from the following shortcomings: 
Firstly, current work on activation functions is dedicated to augmenting ReLUs with a negative, learnable, or context-dependent slope~\cite{ICMLW13LeakyReLU,ICCV15PReLU,Sha2019Modulator}, self-normalizing properties~\cite{NIPS17SELU}, or learnable spatial connections~\cite{CVPR18xUnit}.
However, recent research on neural architecture search~\cite{ICLRW18Swish} shows that activation functions, which abandon the form of ReLU, tend to work better on deeper models across many challenging datasets. 
Secondly, current learnable activation units either impose a global slope on the whole feature map~\cite{ICCV15PReLU,Sha2019Modulator} or aggregate feature context from a large spatial scope~\cite{CVPR18xUnit}. 
However, very different from the attention mechanism favoring the global context, activation functions have a significant inclination on the local context and point-wise gating manner (also see \cref{subsec:ablation}).
Thirdly, the ReLU remains the default activation function for deep neural networks since none of its hand-designed alternatives has managed to show consistent performance improvements across different models and datasets.

In this work, we propose the \emph{attentional activation}~(ATAC) unit to tackle the above shortcomings, which depicts a novel dynamic and context-aware activation function. 
One of our key observation is that both the attention mechanism and the concept of a activation function can be formulated as a non-linear adaptive gating function (see \cref{subsec:unification}). 
More precisely, the activation unit is a non-context aware attention module, while the attention mechanism can be seen as a context-aware activation function.
Besides introducing non-linearity, our ATAC units enable networks to conduct a layer-wise context-aware feature refinement:
\begin{compactenum}
  \item The ATAC units differ from the standard layout of ReLUs and offer a generalized approach to unify the concepts of activation functions and attention mechanisms under the same framework of non-linear gating functions. 
  \item To meet both, the locality of activation functions and the contextual aggregation of attention mechanisms, we propose a local channel attention module, which aggregates point-wise cross-channel feature contextual information. 
  \item The ATAC units make it possible to construct fully attentional networks that perform significantly better with a modest number of additional parameters. 
\end{compactenum}
We conduct extensive ablation studies to investigate the importance of locality and the efficiency of attentional activation as well as fully attentional networks.
To demonstrate the effectiveness of our ATAC units, we also compare them with other activation functions and state-of-the-art attention mechanisms.  
Our experimental results indicate that, given a comparable number of parameters, the models based on our ATAC units outperform other state-of-the-art networks on the well-known CIFAR-10, CIFAR-100, and ImageNet datasets.
\footnote{All the source code and trained models are made publicly available at \url{https://github.com/YimianDai/open-atac}.}

\section{Related Work}
We start by revisiting both activation units as well as modern attention mechanisms used in the context of deep learning.

\subsection{Activation Units} \label{subsec:activation}
%
Activation functions are an integral part of neural networks.
Given finite networks, a better activation function improves convergence and might yield a superior performance.  
The ReLU has been a breakthrough that has alleviated the vanishing gradient problem and has played an important role for the training of the first ``deep'' neural network instances~\cite{ICML10ReLU}. 
However, it also suffers from the so-called ``dying ReLU'' problem, which describes the phenomenon that a weight might become zero and that it does not recover from this state anymore during training.
To alleviate this problem, instead of assigning zero to all negative inputs, the \emph{leaky rectified linear unit} (Leaky ReLU) assigns $\alpha x$ to all negative inputs values $x$, where the slope parameter $\alpha$ (e.g. $\alpha=0.1$) is a hyper-parameter~\cite{ICMLW13LeakyReLU}.
The \emph{gaussian error linear unit}~(GELU) weighs inputs by their magnitude in a non-linear manner, instead of gating inputs by their sign as it is done by ReLUs~\cite{arXiv16GELU}.
Another variant is the \emph{scaled exponential linear unit}~(SELU), which induces self-normalizing properties to enable a high-level abstract representation.
In contrast to such hand-designed units, Ramachandran \textit{et al}.\ have been leveraging automatic search techniques and propose the so-called \emph{Swish} activation function~\cite{ICLRW18Swish}, which resorts to element-wise gating as non-linearity (i.e. $x' = x \cdot \sigma(x)$, where $\sigma$ is the Sigmoid function) and which can be seen as a special non-context aware case of the ATAC unit proposed in this work.

Another way to extend ReLUs is to introduce learnable parameters. 
The \emph{parametric rectified linear unit}~(PReLU) learns the optimal slope for the negative region during training~\cite{ICCV15PReLU}. 
Yang \textit{et al}.\ and Agostinelli \textit{et al}.\ have generalized the form to a learnable piece-wise linear function formulated as a parametrized sum of hinge-shaped functions~\cite{NIPS2016ADMMNet,Agostinelli2015ICLR}.
Recently, Kligvasser \textit{et al.}\ proposed the so-called \emph{xUnit}, a learnable non-linear function that augments ReLU with spatial connections. 
Our ATAC unit follows the idea of learnable activation functions, but differs in at least two important aspects: Firstly, we abandon the idea of improving and augmenting ReLUs as most works do~\cite{ICMLW13LeakyReLU,arXiv16GELU,NIPS17SELU,ICCV15PReLU,Sha2019Modulator,CVPR18xUnit}. 
Instead, we adopt a lightweight and more flexible attention mechanism as activation function.
Secondly, we emphasize the importance of locality in ATAC units, i.e., 
not only the feature context should be locally aggregated, but the attentional weights should also be applied in a point-wise and individual manner.

\subsection{Attention Mechanisms} \label{subsec:attention}
Motivated by their success in natural language processing tasks, attention mechanisms have also been widely employed in computer vision applications~\cite{CVPR18SENet,NIPS18GENet}. 
Most of them are implemented as one or more ``pluggable'' modules that are integrated into the middle or final blocks of existing networks.
The key idea is to adaptively recalibrate feature maps using weights that are dynamically generated via context aggregation on feature maps.   
The \emph{squeeze-and-excitation network}~(SENet)~\cite{CVPR18SENet}, the last winner of the popular ImageNet challenge~\cite{IJCV15ImageNet}, re-weighs channel-wise feature responses by explicitly modeling inter-dependencies between channels via a bottleneck fully-connected (FC) module. 
From the perspective of this work, we refer to SENet as the global channel attention module since it aggregates global context via global averaging pooling (GAP) from entire feature maps.
A popular strategy for improving the performance of attention modules is based on incorporating long-range spatial interactions. 
Woo \textit{et al.}\ propose the so-called \emph{convolutional block attention module} to sequentially apply channel and spatial attention modules to learn ``what'' and ``where'' to focus~\cite{ECCV18CBAM}.
The \emph{gather-excite network}~(GENet) efficiently aggregates feature responses from a large spatial extent by depth-wise convolution~(DWConv) and redistributes the pooled information to local features~\cite{NIPS18GENet}.
To capture long-range interactions, \emph{attention augmented convolutional neural networks}~\cite{ICCV19AugmentedConv} replace convolutions by a two-dimensional relative self-attention mechanism as a stand-alone computational primitive for image classification.

In \cref{Tab:Attention}, we provide a brief summary of related feature context aggregation schemes, in which PWConv denotes the point-wise convolution~\cite{ICLR14NiN} and ShrinkChannel means the global averaging pooling along the channel axis. 
In contrast to the works sketched above, which aim at refining feature maps based on a wide range of contextual information, the proposed local channel attention module emphasizes point-wise channel-wise context aggregation, which we think is a key ingredient to achieve a better activation performance.
Another difference of the proposed ATAC unit is that it goes beyond block/module-level refinement to a layer-wise attentional modulation within activation functions enabling us to build fully attentional networks.

\setlength{\tabcolsep}{2pt}
\begin{table}[t]
\begin{center}
\caption{Context aggregation schemes in attention modules}
\label{tab:schemes}
\label{Tab:Attention}
\footnotesize
\begin{tabular}{Sl Sc Sc Sc} 
\toprule  
Scale       & Interaction  & Formulation & Reference \\
\midrule 
\multirow{2}{*}{Global} & Spatial           & $\mathrm{Conv}(\mathrm{ShrinkChannel}(X))$ & \cite{CVPR17SCACNN,ECCV18CBAM} \\
                        & Channel-wise           & $\text{FC}(\mathrm{GAP}(X))$ & \cite{CVPR18SENet,CVPR19SKNet} \\
\midrule                         
\multirow{1}{*}{Spatial Scope}  & Spatial           & $\text{DWConv}(X)$ & \cite{CVPR18xUnit,NIPS18GENet} \\
\midrule 
Point-wise & Channel-wise & $\text{PWConv}(X)$ & \textbf{\textit{ours}} \\             

\bottomrule
\end{tabular}
\end{center}
\vspace{-\baselineskip}
\end{table}

We notice that concurrent with our work, Ramachandran \textit{et al.}\ developed a local self-attention layer to replace spatial convolutions, which also provides a way to build a fully attentional models~\cite{NIPS19AttentionConv}.
Their experimental analysis indicates that using self-attention in the initial layers of a convolutional network
yields worse results compared to using the convolution stem. 
However, our ATAC units do not suffer from this problem. 
More precisely, in terms of increased parameters, using ATAC units in the initial layers is the most cost-effective solution (see \cref{subsec:ablation}).
We think that the differences can be explained by the different usage of the attention mechanism.
In \cite{NIPS19AttentionConv}, the self-attention layer is used to learn useful features, while the raw pixels at the stem layer are individually uninformative and heavily spatially correlated, which is difficult for content-based mechanisms.
In contrast, our ATAC units are only responsible for activating and refining the features extracted by convolution.

\section{Attentional Activation} \label{sec:atac}
The ATAC units proposed in this work depict a unification of activation functions and attention mechanisms. 
Below, we first provide a unified framework for these two concepts. 
Afterwards, we describe the unit followed by a description of how to integrate it into common neural network architectures.

\subsection{Unification of Attention and Activation} \label{subsec:unification}

Given an intermediate feature map $\mathbf{X} \in \mathbb{R}^{C \times H \times W}$ with $C$ channels and feature maps of size $H \times W$, the transformation induced by attention mechanisms can be expressed as
\begin{equation}
  \mathbf{X}^{\prime}=\mathbf{G}\left(\mathbf{X}\right) \otimes \mathbf{X}, 
  \label{Eq:OverallAttention}
\end{equation}
where $\otimes$ denotes the element-wise multiplication and where $\mathbf{G}(\mathbf{X}) \in \mathbb{R}^{C \times H \times W}$ is a three-dimensional weight map generated by the corresponding attention gating module~$\mathbf{G}$. 
Here, the output of the gating module depends on the whole feature map $\mathbf{X}$. Thus, given a specific position $(c, i, j)$, one can rewrite~\cref{Eq:OverallAttention} in a scalar form as
\begin{align}
  \mathbf{X}^{\prime}_{[c, i, j]} = \mathbf{G}(\mathbf{X})_{[c, i, j]} \cdot \mathbf{X}_{[c, i, j]}
                                  = g_{c, i, j}\left(\mathbf{X}\right) \cdot \mathbf{X}_{[c, i, j]}, 
  \label{Eq:ElemAttention}
\end{align}
where $g$ is a complex gating function. 
Note that, given a position $(c, i, j)$, the function~$g$ is responsible for aggregating the relevant feature context, generating attention weights, and broadcasting as well as picking weights for~$\mathbf{X}_{[c, i, j]}$.

Meanwhile, an activation function can also be formulated in the form of a gating function~\cite{CVPR18xUnit} in the following way:
\begin{align}
  \mathbf{X}^{\prime}_{[c, i, j]} & = g \left(\mathbf{X}_{[c, i, j]} \right) \cdot \mathbf{X}_{[c, i, j]}
  \label{eq:activation}
\end{align}
For instance, given the ReLU activation function, the scalar function $g$ in \cref{eq:activation} is an indicator function. 
For the Swish activation function, $g$ is a Sigmoid function.
For the \emph{sinusoidal representation network} (SIREN)~\cite{arXiv20SIREN} unit, $g$ is the Sinc function $\sin(x)/x$.
Other activation functions follow this formulation in a  similar way.

Comparing \cref{Eq:ElemAttention} and \cref{eq:activation}, it can be seen that both the attention mechanism and the concept of activation functions give rise to non-linear adaptive gating functions. 
Despite the specific forms, their only difference is that the activation gating function $g$ takes a scalar as input and outputs a scalar, whereas the attention gating function $g_{c,i,j}$ takes a larger spatial, global, or cross-channel feature context as input.
This connection between activation functions and attention mechanisms motivates the use of lightweight attention modules as activation functions.
Besides introducing non-linearity, such attentional activation units enable the networks to conduct an adaptive layer-wise context-aware feature refinement.

\subsection{Local Channel Attention Module} \label{subsec:chaatac}
\begin{figure}[t]
    \centering
    \includegraphics[height=10\baselineskip]{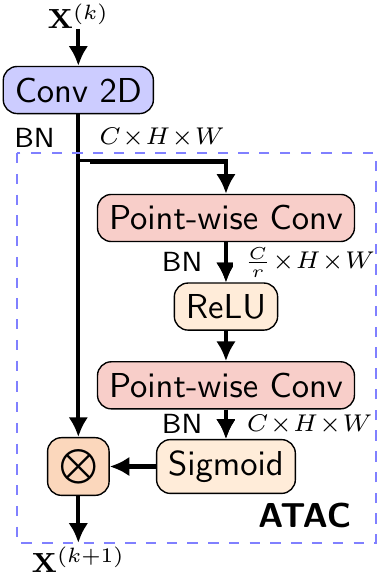}
    \caption{The proposed attentional activation (ATAC) unit}
    \label{fig:ChaATAC}
    \vspace{-1\baselineskip}
\end{figure}

In this work, we use attention modules as activation functions throughout the entire network architecture.
Therefore, the attentional activation functions must be cheap from a computational perspective.
To simultaneously satisfy the locality requirements of activation functions and the contextual aggregation requirements of attention mechanisms, our ATAC units resort to point-wise convolutions~\cite{ICLR14NiN} to realize local attention, which are a perfect fit since they map cross-channel correlations in a point-wise manner and also exhibit a few parameters.

The architecture of the proposed local channel attention based attentional activation unit is illustrated in \cref{fig:ChaATAC}:
The goal is to enable the network to selectively and element-wisely activate and refine the features according to the point-wise cross-channel correlations. 
To save parameters, the attentional weight $\mathbf{L}(\mathbf{X}) \in \mathbb{R}^{C \times H \times W}$ is computed via a bottleneck structure as follows:
\begin{equation}
\mathbf{L}(\mathbf{X}) = \sigma\left( \mathcal{B} \left(\mathrm{PWConv}_2 \left(\delta\left( \mathcal{B} \left(\mathrm{PWConv}_1 (\mathbf{X}) \right)\right)\right)\right) \right)
\end{equation}
Here, $\delta$ is the ReLU activation function and $\mathcal{B}$ denotes the batch normalization operator~\cite{ICML15BN}. 
The kernel size of $\mathrm{PWConv}_1$ is $\frac{C}{r} \times C \times 1 \times 1$ and the kernel size of $\mathrm{PWConv}_2$ is $C \times \frac{C}{r} \times 1 \times 1$. 
The parameter $r$ is the channel reduction ratio.
It is noteworthy that $\mathbf{L}(\mathbf{X})$ has the same shape as the input feature maps and can thus be used to activate and highlight the subtle details in a local manner---spatially and across channels.
Finally, the activated feature map $\mathbf{X}^{\prime}$ is obtained via an element-wise multiplication with $\mathbf{L}(\mathbf{X})$:
\begin{equation}
  \mathbf{X}^{\prime} = \mathbf{L}(\mathbf{X}) \otimes \mathbf{X}
\end{equation}

\begin{figure}[t]
    \centering
    \subfloat[]{
        \includegraphics[height=12\baselineskip]{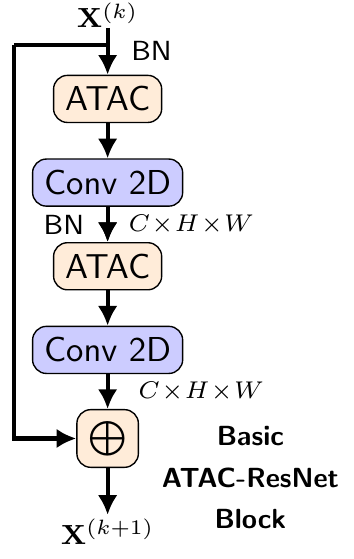}\label{subfig:basic}
    } 
    \hspace{0.5cm}
    \subfloat[]{
        \includegraphics[height=12\baselineskip]{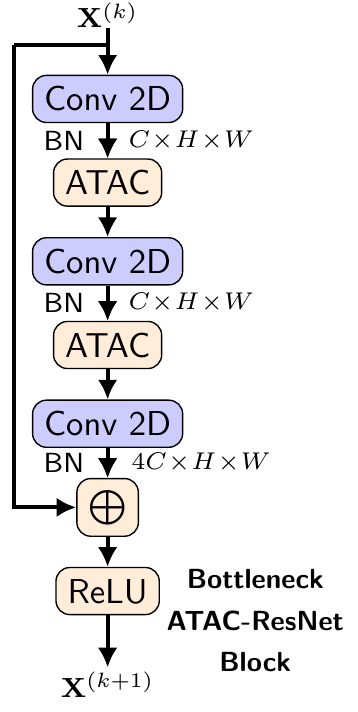}\label{subfig:bottleneck}
    }                   
    \caption{Illustrations of the proposed architectures: (a) The basic ATAC-ResNet-V2 block and (b) the bottleneck ATAC-ResNet-V1b block.
    These blocks can be used to create fully attentional networks by replacing every ReLU with an ATAC unit in a baseline convolutional network.
    }
    \vspace{-1\baselineskip}
\end{figure}

\subsection{Fully-Attentional Networks} \label{subsec:fully}
The unit described above can be used to obtain fully attentional neural networks by replacing every ReLU with an ATAC unit.
Due to this paradigm, we can achieve feature refinement at very early stages, even after the first convolutional layer.
In comparison with a $3\times 3$ convolution, our ATAC unit induces about $\frac{2}{9r}$ additional parameters and computations.  

Our experimental evaluation shows that it is worth spending these additional memory and computing resources for the ATAC units instead of, e.g., making the networks deeper. 
That is, instead of simply increasing the depth of a network, one should pay more attention to the quality of the feature activations.
We hypothesize that the reason behind this is that suppressing irrelevant low-level features and prioritizing relevant object features at earlier stages enables networks to encode higher-level semantics more efficiently.

In the experimental evaluation provided below, we consider the ResNet family of architectures as host networks. 
More specifically, we replace the ReLU unit in the basic ResNet block and bottleneck ResNet block, which we call ATAC-ResNet blocks.
This induced architectures are shown in \cref{subfig:basic} and \cref{subfig:bottleneck} and the details are provided in \cref{Tab:Backbone}.

\newcommand{\blockb}[2]{$
\begin{bmatrix}
\begin{array}{l}
    \!\!\!\!3\!\times\!3 \mathrm{~conv}, #1\!\!\!\!\!\\
    \!\!\!\!3\!\times\!3 \mathrm{~conv}, #1\!\!\!\!\!\\    
\end{array}
\end{bmatrix}\!\!\times\!#2$}

\newcommand{\blockx}[3]{$
\begin{bmatrix}
\begin{array}{l}
    \!\!\!\!3\!\times\!3 \mathrm{~conv}, #1\!\!\!\!\!\\
    \!\!\!\!3\!\times\!3 \mathrm{~conv}, #1\!\!\!\!\!\\ 
    \!\!\!\!3\!\times\!3 \mathrm{~conv}, #2\!\!\!\!\!\\ 
\end{array}
\end{bmatrix}\!\!\times\!#3$}

\newcolumntype{x}[1]{>\centering p{#1pt}}
\newcommand{\ft}[1]{\fontsize{#1pt}{1em}\selectfont}
\renewcommand\arraystretch{1.25}
\setlength{\tabcolsep}{2pt}

\begin{table}[t]
\begin{center}
\begin{tabular}{Sc Sc Sc Sc Sc }
\toprule  
Stage & Output & ResNet-20 & Output & ResNet-50 \\
\midrule 
conv1  & $32\!\times\!32$  & \hspace{-2.25em} $3\!\times3~\mathrm{conv}, 16$ & $112\!\times\!112$ & \hspace{-2.25em} $7\!\times7~\mathrm{conv}, 64$ \\ 
stage1 & $32\!\times\!32$ & \blockb{16}{b}  & $112\!\times\!112$ & \blockx{64}{256}{3} \\
stage2 & $16\!\times\!16$ & \blockb{32}{b}  & $56\!\times\!56$ & \blockx{128}{512}{4} \\
stage3 & $8\!\times\!8$   & \blockb{64}{b}  & $28\!\times\!28$ & \blockx{256}{1024}{6} \\
stage4 &                  &                 & $14\!\times\!14$ & \blockx{512}{2048}{3} \\
\cmidrule{2-5}
& $1\!\times\!1$ & \multicolumn{3}{c}{Averge Pool, 10/100/1000-d FC, Softmax}      \\  
\bottomrule
\end{tabular}
\end{center}
\caption{The host network architectures with ReLUs being replaced by the ATAC units.
ResNet-20 \cite{ECCV16ResNetV2} is used for the CIFAR-10/100 datasets and ResNet-50 \cite{CVPR16ResNetV1} is used for the ImageNet dataset.
For ResNet-20, we scale the model depth using different assignments for the parameter $b$ (which defines the number of blocks per stage) to study the relationship between network depth and the induced performance. Note that $b = 3$ corresponds to the standard ResNet-20 backbone.}
\label{Tab:Backbone}
\vspace{-1\baselineskip}
\end{table}



\section{Experiments}\label{sec:experiment}

To analyze the potential of the proposed ATAC units, we consider several datasets and compare the architectures mentioned above with state-of-the-art baselines. 
We also conduct a comprehensive ablation study to investigate the effectiveness of the design of the ATAC units and the behavior of the networks that are induced by them. 
In particular, the following questions will be investigated in our experimental evaluation:
\begin{compactenum}
    \item Q1: Generally, the attention mechanisms in vision tasks capture long-range contextual interactions to address the semantic ambiguity. Our attentional activation utilizes point-wise local channel attention instead. In our study (see \cref{tab:ablation}), we investigate the question of how important the local context locality is for ATAC units.
    \item Q2: Using a micro-module to enhance the network discriminability is not a new idea. The \emph{network-in-network}~(NiN)~\cite{ICLR14NiN} approach, attentional feature refinement (SENet)~\cite{CVPR18SENet}, and the proposed attentional activation all fall into this idea.  
    Given the same increased computation and parameter budget, we investigate the question if the proposed ATAC units depict a better alternative compared to the NiN-style block, which deepens the network, and attention module like SENet~\cite{CVPR18SENet}, which refines the feature maps (see \cref{tab:ablation}).
    \item Q3: A natural question is also if the performance will improve consistently when more and more ReLUs are replaced by ATAC units until a fully attentional network is obtained. We will examine this as well if ATAC units shall be used in the initial layers of a convolutional network,
    see \cref{fig:comparsion}.
    \item Q4: Next, we will analyze how the networks with our ATAC units compare to networks with other activation functions and other state-of-the-art attention mechanisms. In particular, considering the fact that the proposed ATAC units also induce additional parameters and computational costs compared to their non-parametric alternatives such as ReLU or Swish, we will investigate if convolutional networks with ATAC units yield a superior performance with fewer layers and parameters, see \cref{fig:comparsion} and \cref{Tab:ImageNet}.
    \item Q5: Finally, we will examine if deeper network such as ResNet-50 with our ATAC units suffer from the vanishing gradient problem induced by the Sigmoid function, see \cref{subsec:sota}.
\end{compactenum}

\subsection{Experimental Settings}
For experimental evaluation, we resort to the CIFAR-10, CIFAR-100~\cite{Krizhevsky09CIFAR}, and ImageNet~\cite{IJCV15ImageNet} datasets.
All network architectures in this work are implemented using the MXNet~\cite{NIPSW15MXNet} framework. 
Since most of the experimental architectures cannot take advantage of pre-trained weights, every architecture instantiation is trained from scratch for fairness.
The strategy described by He \textit{et al.}~\cite{ICCV15PReLU} is used for weight initialization.
and the channel reduction ratio $r$ is set to $2$ for all experiments.
For CIFAR-10 and CIFAR-100, the ResNet-20v2~\cite{ECCV16ResNetV2} architecture was adopted as a host backbone network with a single $3\times 3$ convolutional layer, followed by three stages each having three basic residual blocks.
To study the network's behavior with the proposed ATAC units under different computational and parameter budgets, we vary the depths of the models using the block number $b$ in each stage in \cref{Tab:Backbone}.
The Nesterov accelerated SGD (NAG) optimizer was considered for training the models using a learning rate of 0.2, a total number of 400 epochs, weight decay of 1e-4, batch size of 128, and learning rate decay factor of 0.1 at epochs 300 and 350, respectively.
For ImageNet, we resorted to the ResNet-50-v1b~\cite{CVPR16ResNetV1} architecture as a host network and NAG as optimizer with a learning rate of 0.075, a total number of 160 epochs, no weight decay, a batch size of 256 on two GPUs, and a decay of the learning rate by a factor of 0.1 at epoch 40 and 60, respectively.
To save computational time, only the ReLUs in the last two stages of ResNet-50-v1b are replaced with our ATAC units illustrated in \cref{subfig:bottleneck}.

\subsection{Ablation Study}  \label{subsec:ablation}
We start by investigating the questions Q1-Q3 raised above. 
In particular, we consider several micro-module competitors and compare their performances with our approach that replaces each ReLU by an ATAC unit (ATAC(ours) in \cref{tab:ablation}). 
For these experiments, we consider the CIFAR-10 and CIFAR-100 datasets and the ResNet-20 as host network with a varying number $b$ of blocks.

\subsubsection{Importance of Locality (Q1)} \label{subsubsec:locality}
\begin{wrapfigure}{r}{3.1cm}
    \vspace{-25pt}
    \centering
     \resizebox{3.0cm}{!}{
      \includegraphics[height=0.27\textwidth]{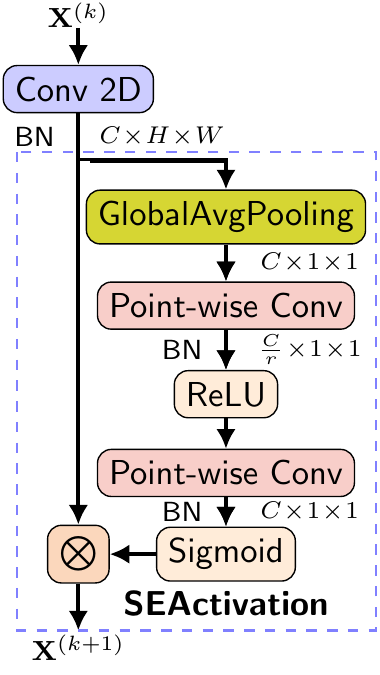}
      }
    \vspace{-15pt}
    \caption{SEActivation
\label{fig:SEActivation}
}
\vspace{-12pt}
\end{wrapfigure}

We start by comparing the ATAC unit with the SEActivation unit shown in Figure~\ref{fig:SEActivation}, which corresponds to the SE block used by the SENet~\cite{CVPR18SENet}. Note that, instead of using this block for block-wise feature refinement, we make use of this unit as layer-wise activation, i.e., as replacement for the ReLU.
Compared to our ATAC unit, the SEActivation unit adds global average pooling layer at the beginning to obtain $C$ feature maps of size $1 \times 1$.
While the ATAC unit and the SEActivation unit have the same number of parameters, they vary w.r.t.\ 
the contextual aggregation scale and the application scope of the attentional weight. 
More precisely, 
SEActivation aggregates the global contextual information and each feature map of size $H \times W$ shares the same attentional activation weight of size $1 \times 1$ in the end. 
In contrast, the ATAC unit captures the channel-wise relationship in a point-wise manner and each scalar in the feature map has an individual gating weight (i.e.\ the weight tensor has size $C \times H \times W$). 

\cref{tab:ablation} presents the comparison between both units on CIFAR-10 and CIFAR-100 given a gradually increased network depth. 
It can be seen that, compared with ATAC, the performance of the network using the SEActivation unit is significantly worse (note that the block is used in a different fashion here compared to its original usage as feature refinement in the SENet architecture).
The results suggest that the locality is of vital importance in case such units are used for attentional activation.

\setlength{\tabcolsep}{1.5pt}
\begin{table}[t]
\begin{center}
\caption{
Comparision of the classification accuracies on CIFAR-10 and CIFAR-100. 
The results comparing SEActivation and ATAC suggest that the locality of contextual aggregation is of vital importance.
The comparison among NiN, LocalSENet, and ATAC suggests that given the same computational and parameter budget, the layer-wise feature refinement via the proposed attentional activation outperforms going deeper via a NiN-style block and block-wise feature refinement via attention module.
}
\label{tab:ablation}
\resizebox{1.0\columnwidth}{!}{
\begin{tabular}{r c c c c c c c c} 
\toprule  
Scale                            & \multicolumn{4}{c}{CIFAR-10} & \multicolumn{4}{c}{CIFAR-100} \\
                    \cmidrule(lr){2-5}                           \cmidrule(lr){6-9}
              & $b=1$ & $b=2$ & $b=3$ & $b=4$ & $b=1$ & $b=2$ & $b=3$ & $b=4$  \\
\midrule
ReLU & 0.895 & 0.920 & 0.929 & 0.935 & 0.737 & 0.785 & 0.799 & 0.806 \\
SEActivation & 0.548 & 0.601 & 0.613 & 0.622 & 0.388 & 0.432 & 0.452 & 0.456           \\
\addlinespace
NiN        & 0.893 & 0.917 & 0.922 & 0.926 & 0.743 & 0.776 & 0.792 & 0.796           \\
LocalSENet    & 0.906 & 0.926 & 0.931 & 0.937 & 0.762 & 0.794 & 0.805 & 0.811           \\
\addlinespace
ATAC (\textit{\textbf{ours}}) & \textbf{0.906} & \textbf{0.927} & \textbf{0.936} & \textbf{0.939} & \textbf{0.764} & \textbf{0.796} & \textbf{0.812} & \textbf{0.821}  \\
\bottomrule
\end{tabular}
}
\end{center}
\vspace{-2\baselineskip}
\end{table}

\subsubsection{Activation vs. NiN vs. Refinement (Q2)}
\begin{wrapfigure}{r}{2.7cm}
    \vspace{-22pt}
    \centering
     \resizebox{2.4cm}{!}{
      \includegraphics[height=0.27\textwidth]{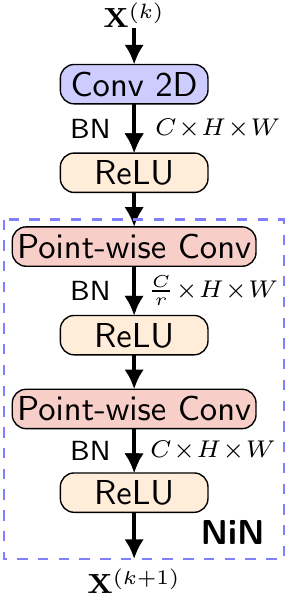}
      }
    \vspace{-0pt}
    \caption{NiN
\label{fig:NiN}
}
\vspace{-10pt}
\end{wrapfigure}

Next, we investigate and compare our ATAC units with two other micro-modules: The first one is a NiN-style block~\cite{ICLR14NiN}, which introduces point-wise convolutions after convolution layers to enhance the network discriminability. \cref{fig:NiN} provides an instantiation of the NiN-style block, which has an increased number of parameters compared to its original implementation to be on-pair with the the other modules. The NiN-module is applied after each convolution layer and the associated ReLU in the ResNet-20 host network.

\begin{wrapfigure}{r}{3.0cm}
    \vspace{-15pt}
    \centering
     \resizebox{2.9cm}{!}{
      \includegraphics[height=0.27\textwidth]{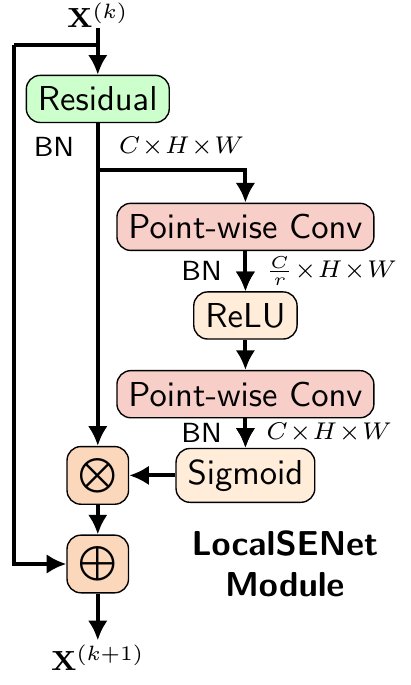}
      }
    \vspace{-15pt}
    \caption{LocalSENet
\label{fig:LocalSENet}
}
\vspace{-5pt}
\end{wrapfigure}
The second micro-module considered here as competitor is the LocalSENet block shown in \cref{fig:LocalSENet}. Basically, instead of using the ATAC unit as replacement for ReLU, the LocalSENet uses the same local attention mechanism to refine the residual output. 
Compared to the SEActivation module, we employ a local channel attention module in LocalSENet. Furthermore, we do not use it as activation function. Thus, the induced networks correspond to the SENet \emph{without} the global average pooling block for the residual branch.
Note that the LocalSENet block refines the residual output after two convolutions, whereas the NiN and ATAC modules are applied after each convolutional layer. 
Hence, to obtain the same number of parameters and the same computational costs, the channel reduction ratio $r$ in the LocalSENet module is set to~1.0.


\cref{tab:ablation} provides the results, from which it can be seen that:
1)~The performance of NiN is not as good as LocalSENet and ATAC, which suggests that having a small additional budget for parameters and computation costs, one should resort to the attention mechanism instead of a NiN-style block. 
This strengthens our assumption that instead of blindly increasing the network depth, refining the feature maps is a more efficient and effective way to increase the networks' performance. 
2)~The difference between LocalSENet and ATAC is that LocalSENet uses the attention mechanism only once with all the additional parameters.
In contrast, the ATAC units are applied after every convolution (with each ATAC unit having only half the parameters compared to the LocalSENet module). 
The results suggest that given the same budget for parameters and computational costs, one should choose the paradigm that applies as many lightweight attention modules as possible, instead of adopting the sophisticated attention modules only a few times. 

\subsubsection{Towards Fully Attentional Networks (Q3)}

\begin{figure}[t]
    \centering
    \subfloat[]{
        \includegraphics[width=0.235\textwidth]{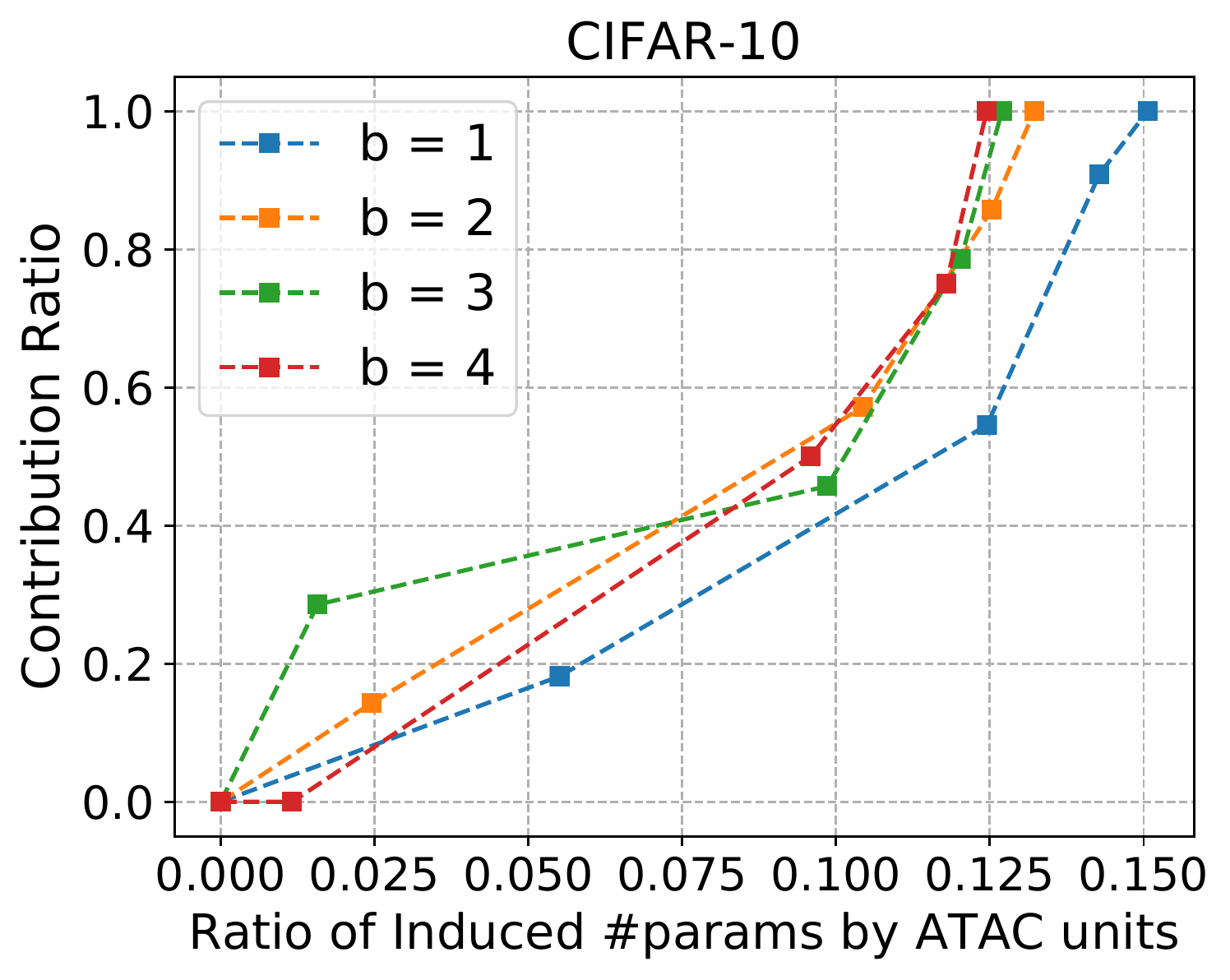}\label{subfig:fully10}
    }\hspace{-1em}       
    \subfloat[]{
        \includegraphics[width=0.235\textwidth]{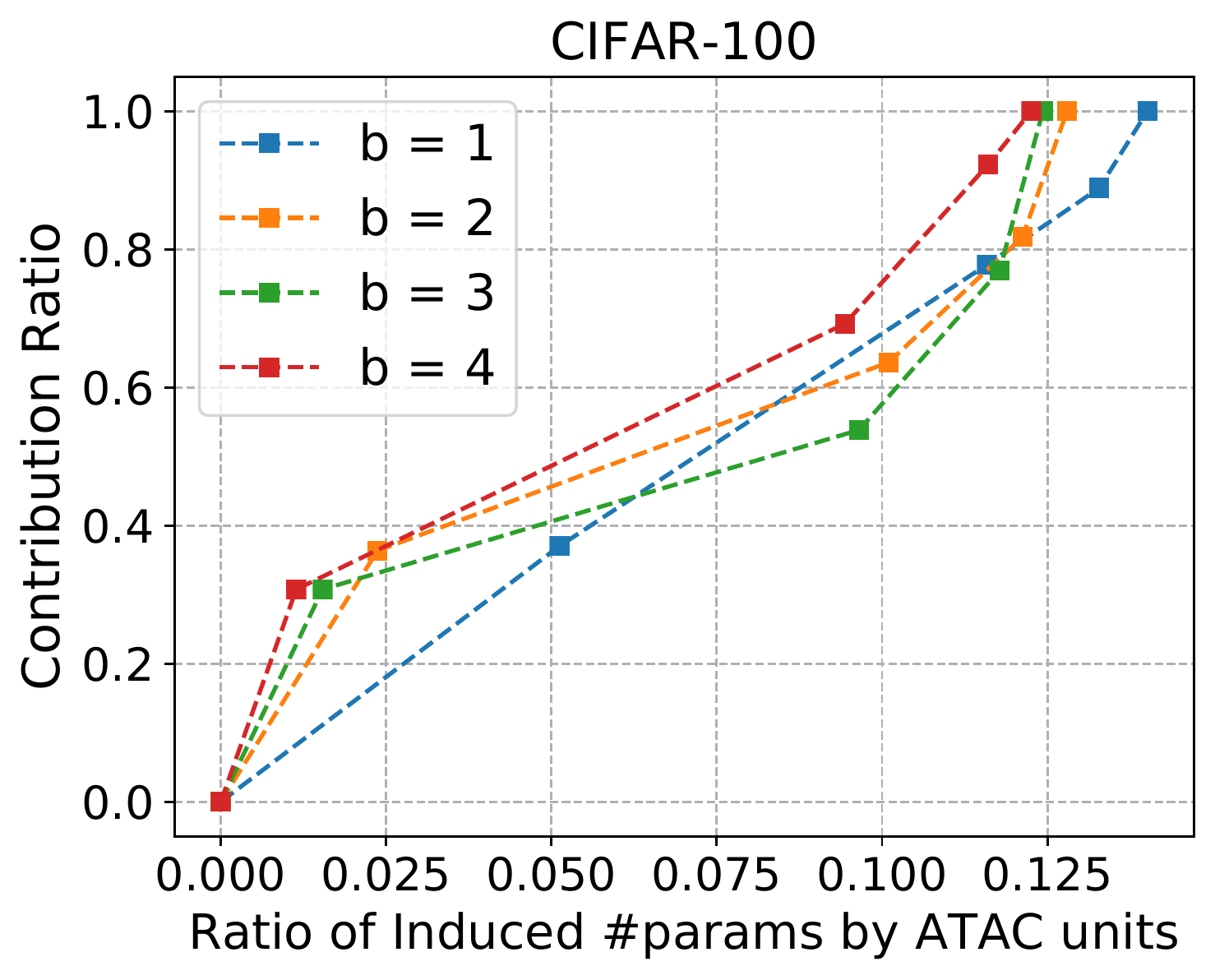}\label{subfig:fully100}
    }           
    \caption{Illustration of the performance gain tendency in percentage by gradually replacing ReLUs with the proposed ATAC units starting with the last layer and ending with the first layer on CIFAR-10 and CIFAR-100. Here, a contribution ratio of 1.0 corresponds to the normalized performance gain obtained via a fully-attentional network with all ReLUs being replaced by ATAC units and a ratio of 0.0 corresponds to no ATAC units being used (hence, no gain).
    The results suggest that the network can obtain a consistent performance gain by going towards a fully attentional network.
    }
    \label{fig:fully}
    \vspace{-1\baselineskip}    
\end{figure}


            
We also investigated the cost-effective ratio of the fully attentional network with the proposed ATAC unit. 
We analyze the network's predictive performance on CIFAR-10 and CIFAR-100 while gradually replacing ReLUs with ATAC units, starting with the last layer and ending with the first layer.
As it can be seen in \cref{subfig:fully10} and \cref{subfig:fully100}, the performance tends to increase with more ATAC units.
Therefore, a fully attentional network offers a way to obtain a performance increase with marginal additional costs.
It can also be seen that the largest performance increase is obtained for the replacements made at the end of the process (steep increase in the range from 0.125 to 0.150), which correspond to replacements of ReLUs be ATAC units in the first layers of the network.
This supports our hypothesis that early attentional modulation enables networks to encode higher-level semantics more efficiently by suppressing irrelevant low-level features and highlighting relevant features in the early layers of the networks.

\subsection{Comparison} \label{subsec:sota}
Finally, we address question Q4 raised at the beginning of this section by comparing our approach with several activation functions and other state-of-the-art network competitors. We also show that the ATAC-ResNet is not affected by the vanishing gradient problem, thus addressing question Q5.

\subsubsection{Activation Units \& Networks (Q4)} 
\begin{figure}[t]
    \centering
    \subfloat[]{
        \includegraphics[width=0.24\textwidth]{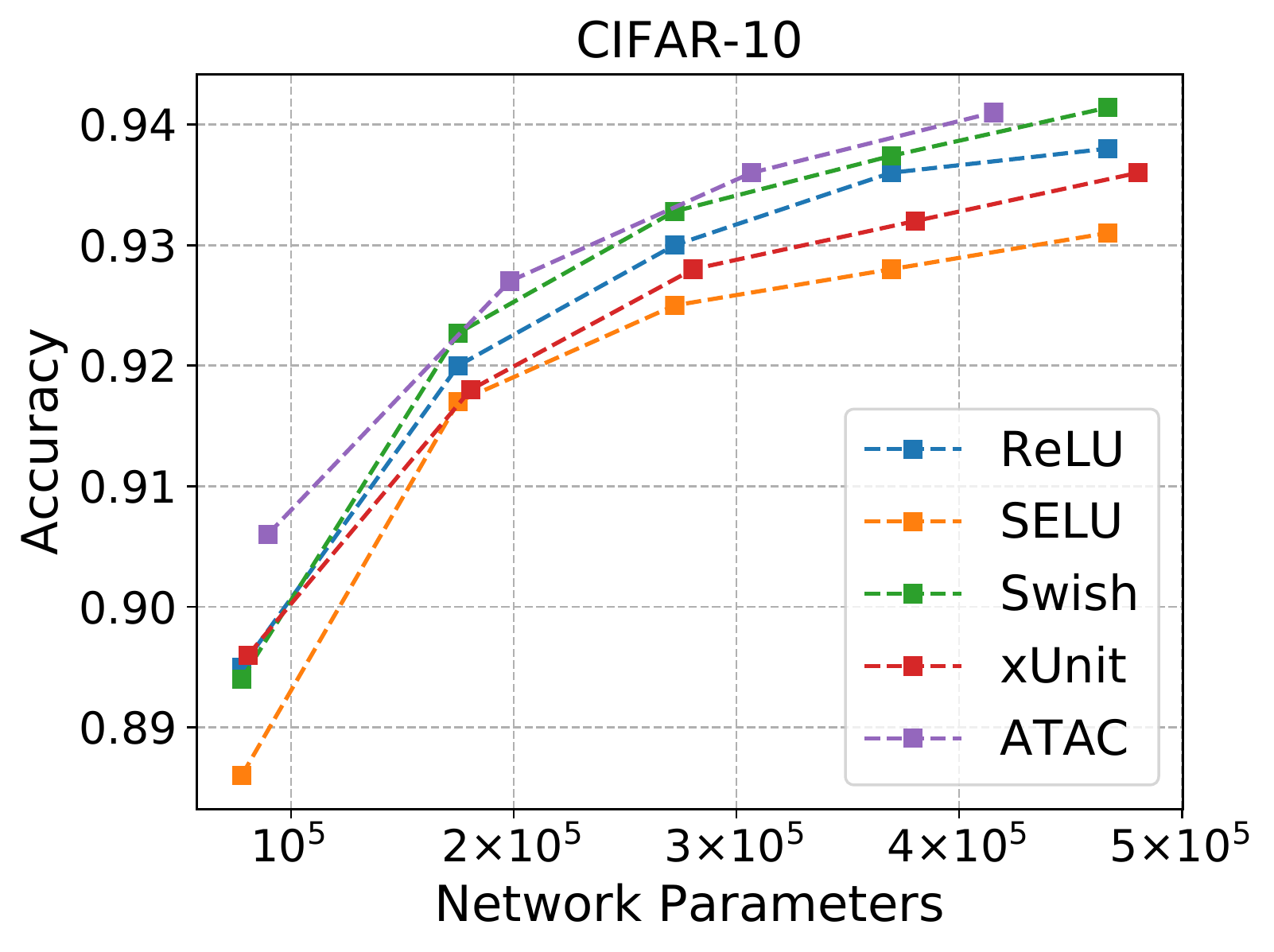}\label{subfig:activation10}
    }\hspace{-1.5em}   
    \subfloat[]{
        \includegraphics[width=0.24\textwidth]{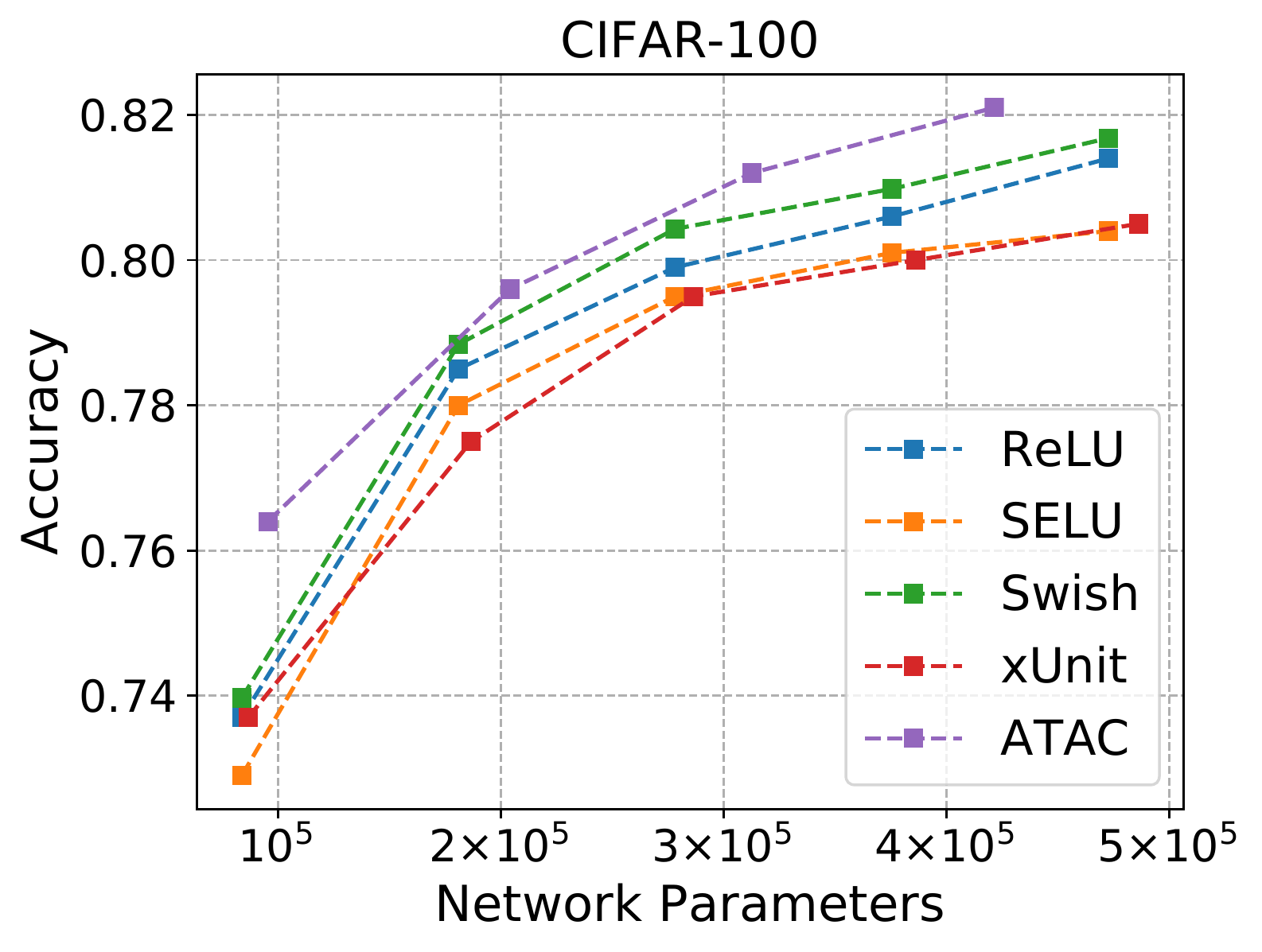}\label{subfig:activation100}
    }    
    \caption{Comparison with other activation units on CIFAR-10 and CIFAR-100 with a gradual increase of network depth.
    The results suggest that we can obtain a better performance with even fewer layers or parameters per network by replacing ReLUs with the proposed ATAC units.
    }
    \label{fig:activation}
    \vspace{-1\baselineskip}    
\end{figure}

First, we compare the proposed ATAC unit with other activation units, namely ReLU~\cite{ICML10ReLU}, SELU~\cite{NIPS17SELU}, Swish~\cite{ICLRW18Swish}, and xUnit~\cite{CVPR18xUnit}.\footnote{We also consdiered GELU~\cite{arXiv16GELU} and PReLU~\cite{ICCV15PReLU}, but their performance was not as good as the aforementioned baselines and were, hence, excluded from the overall comparison.}
\cref{subfig:activation10} and \cref{subfig:activation100} provide the comparison on CIFAR-10 and CIFAR-100 given a gradual increase of the depths of the networks.
It can be seen that:
(a)~The ATAC unit achieves a better performance for all experimental settings, which demonstrates its effectiveness compared to the baselines.
The Swish unit, which is also a non-linear gating function, ranked second in the comparison, better than the ReLU-like activation units.
These results reaffirm that one can obtain better activation functions by considering alternative non-linearities than ReLU-like units.
(b)~Since the ATAC unit outperforms the Swish unit---which can be interpreted as a non-context-aware scalar version of the ATAC unit---we conclude that channel-wise context is beneficial for activation functions.
(c)~By replacing ReLUs by ATAC units, one can obtain a more efficient convolutional network that yields a better performance with fewer layers or parameters per network.
For example, in \cref{subfig:activation100}, the ATAC-ResNet ($b=3$) achieves the same classification accuracy as the ResNet ($b=5$), while only using 65\% of the parameters.

\begin{figure}[t]
    \centering
    \subfloat{
        \includegraphics[width=0.235\textwidth]{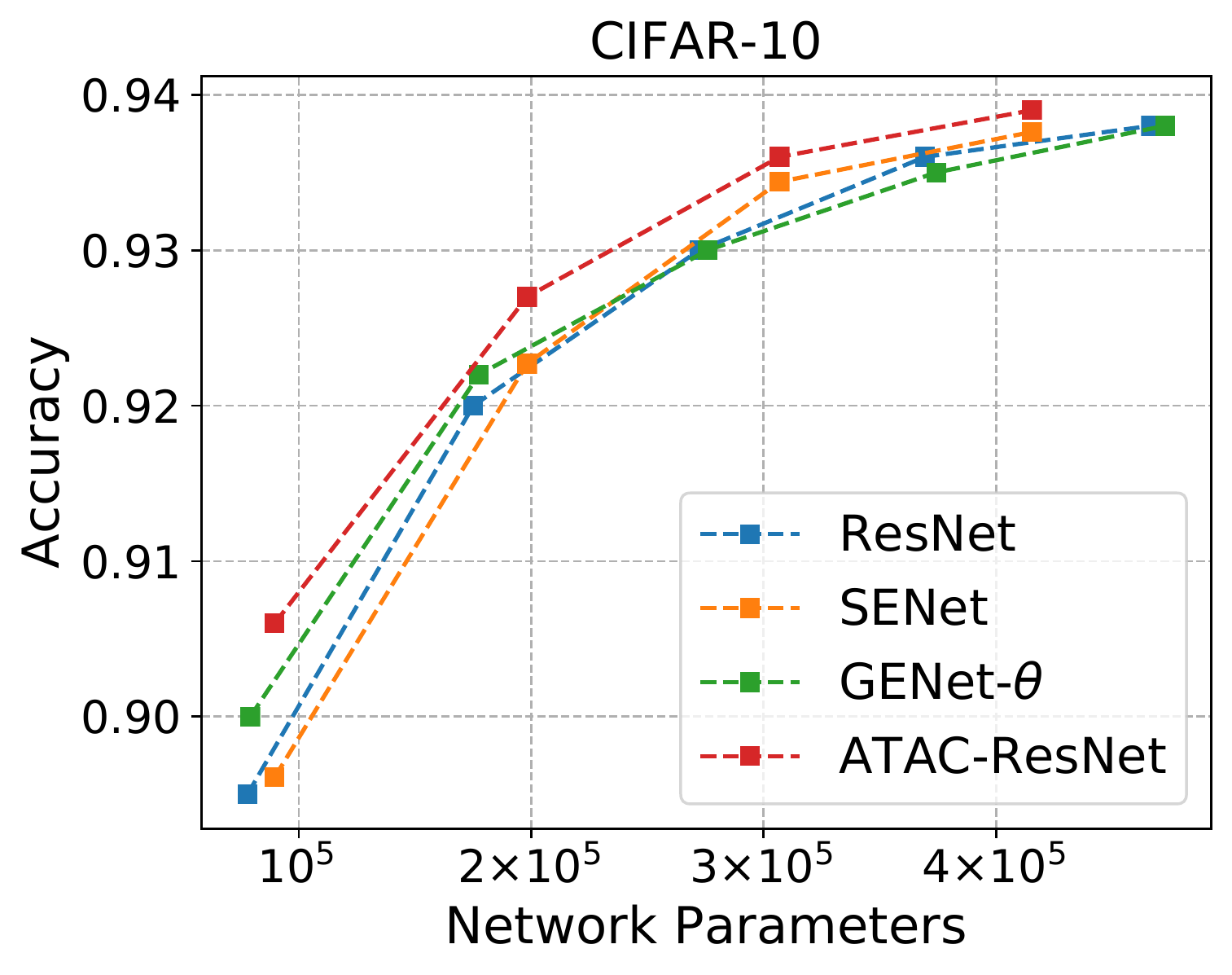}\label{subfig:sota10}
    }\hspace{-1em}    
    \subfloat{
        \includegraphics[width=0.235\textwidth]{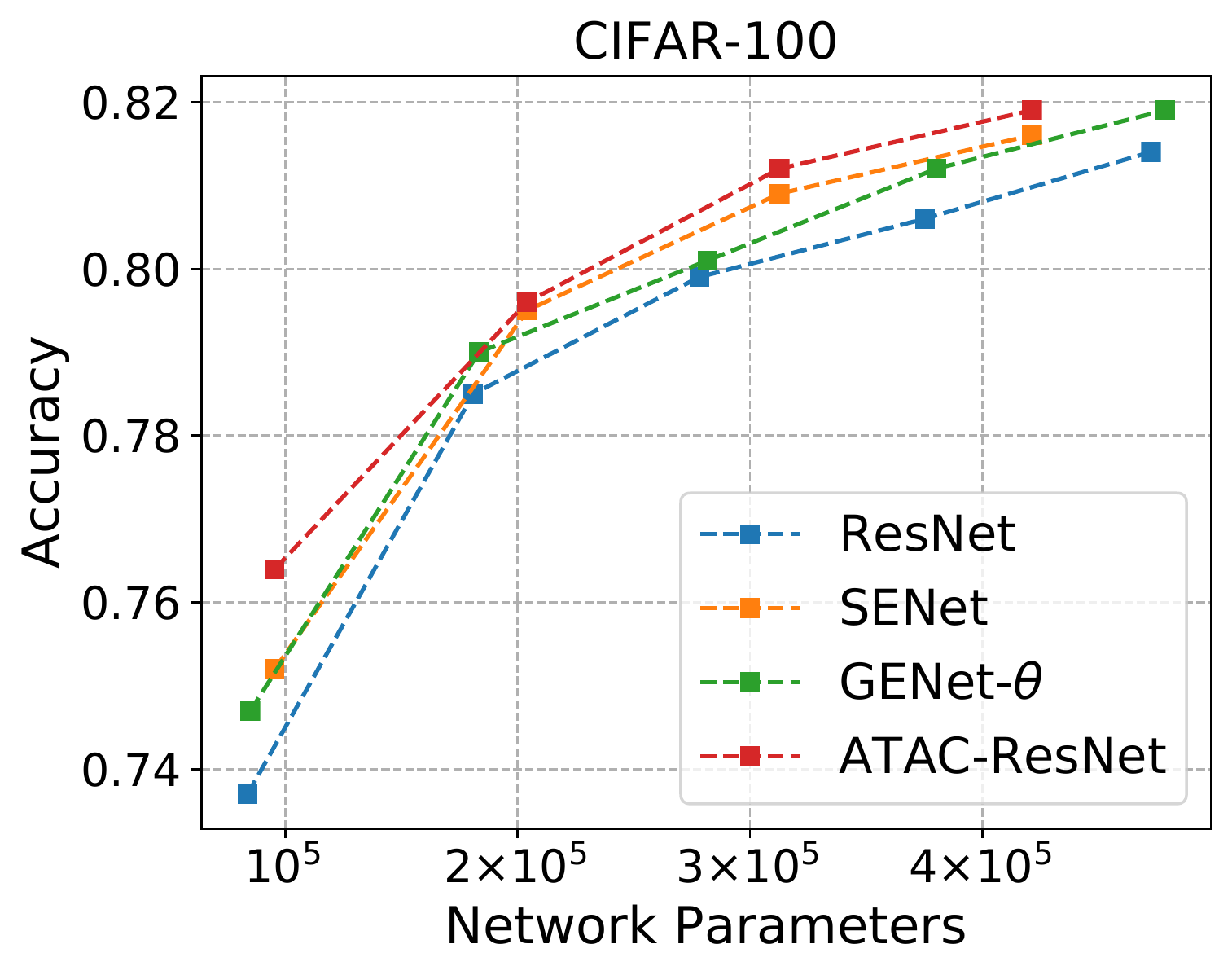}\label{subfig:sota100}
    }  
    \caption{
    Comparison of different networks on CIFAR-10/100 given a gradual increase of network depth.
    The results demonstrate the effectiveness of the layer-wise manner of simultaneous feature  activation and refinement by our ATAC units.
    }
    \label{fig:comparsion}
    \vspace{-1\baselineskip}
\end{figure}

Next, we compared our proposed methods with the baseline and other state-of-the-art networks.
\cref{subfig:sota10} and \cref{subfig:sota100} illustrate the results given a gradual increase in the network depth for all networks on CIFAR-10 and CIFAR-100.
Comparing SENet~\cite{CVPR18SENet} and GENet~\cite{NIPS18GENet}, one can see that the network with the ATAC units performs better for all settings.
We believe that the improved performance stems from the layer-wise and simultaneous feature activation and refinement of the proposed attentional activation scheme. 
We also validated our ATAC units on the ImageNet dataset. 
The results are provided in \cref{Tab:ImageNet}.
Note that compared to the self-reported results from SENets~\cite{CVPR18SENet}, attention augmented (AA) convolution networks~\cite{ICCV19AugmentedConv}, full attention (FA) vision models~\cite{NIPS19AttentionConv}, and gather-excite (GE) networks~\cite{NIPS18GENet}, our ATAC-ResNet-50 achieves the best in top-1 error.
Notably, compared with GE-$\theta^+$-ResNet-50 and SE-ResNet-50, the ATAC-ResNet-50 requires fewer parameters.
Although the number of GFlops is a bit higher (which mainly stems from the point-wise convolutions in the local contextual aggregation), the network yields a good trade-off compared to the other architectures.

\subsubsection{Vanashing Gradients (Q5)}
The performance of our ATAC-ResNet on CIFAR-10/100 ($b=5$, ResNet-32) and ImageNet
(ResNet-50) empirically answers the question Q5 that deep networks equipped with ATAC units do not appear to
suffer from the vanishing gradient problem. 
Interestingly, 
the Swish activation function~\cite{ICLRW18Swish} also adopts the Sigmoid function and the network with Swish units can also go very deep.
Note that the Sigmoid and Softmax functions are not used as activation in our context, but are used to obtain probabilities to weigh the feature maps (which cannot be obtained via ReLUs).
This is the reason why attention mechanism modules, deep belief networks (DBN), recurrent neural networks (RNN), and long short-term memory (LSTM) networks as well as our ATAC units typically have layer-wise Sigmoid or Softmax functions and can go very deep.
In fact, the emergence of batch normalization~\cite{ICML15BN} allows the Sigmoid function to be used in deep networks again. 
With the better expressive ability of Sigmoid, these networks have achieved better results.
In addition, the residual connection \cite{CVPR16ResNetV1} also helps training very deep networks.

\setlength{\tabcolsep}{3pt}
\begin{table}[t]
\begin{center}
\caption{Classification comparison on ImageNet with other state-of-the-art networks. ATAC-ResNet-50 achieves the best top-1 err. with smaller parameter numbers than SE-ResNet-50 and GE-$\theta^+$-ResNet-50.}
\label{Tab:ImageNet}\vspace{-.125cm}
\small
\begin{tabular}{Sr Sl Sc Sc Sc} 
\toprule  
Architecture                              & GFlops & Params & top-1 err.     & top-5 err. \\
\midrule 
ResNet-50   \cite{CVPR16ResNetV1}          & 3.86   & 25.6M  & 23.30          & 6.55 \\
SE-ResNet-50 \cite{CVPR18SENet}            & 3.87   & 28.1M  & 22.12          & 5.99 \\
AA-ResNet-50 \cite{ICCV19AugmentedConv}    & 8.3    & 25.8M  & 22.30          & 6.20 \\
FA-ResNet-50 \cite{NIPS19AttentionConv}    & 7.2    & 18.0M  & 22.40          & /    \\
GE-$\theta^+$-ResNet-50 \cite{NIPS18GENet}   & 3.87   & 33.7M  & 21.88          & 5.80 \\
ATAC-ResNet-50 (\textbf{\textit{ours}}) & 4.4    & 28.0M  & \textbf{21.41} & 6.02 \\
\bottomrule
\end{tabular}
\end{center}
\vspace{-2\baselineskip}
\end{table}




\section{Conclusion}

Smarter activation functions that integrate what is generally considered separately as attention mechanisms are very promising and worthy of further research. 
Instead of blindly increasing the depth of network, one should pay more attention to the quality of feature activation. In
particular, we found that our attentional activation units---a unification of activation function and attention mechanism that endow activation units with attentional context information---improve the performance of all the networks and datasets that we have experimented with so far. 
To meet both the locality of activation function and contextual aggregation of attention mechanism, we propose a local channel attention module, which locally aggregates point-wise cross-channel feature contextual information.
A simple procedure of replacing all ReLUs with the proposed ATAC units produces a fully attentional network that performs significantly better than the baseline with a modest number of additional parameters.
Compared with other activation units, the convolutional network with our ATAC units can gain a performance boost with fewer layers or parameters per network. 


\section*{Acknowledgment}
This work was supported in part by the National Natural Science Foundation of China under Grant No. 61573183, the Open Project Program of the National Laboratory of Pattern Recognition (NLPR) under Grant No. 201900029, the Nanjing University of Aeronautics and Astronautics PhD short-term visiting scholar project under Grant No. 180104DF03, the Excellent Chinese and Foreign Youth Exchange Program, China Association for Science and Technology, China Scholarship Council under Grant No. 201806830039.

\bibliographystyle{IEEEtran}
\bibliography{root}

\end{document}